\documentclass[letterpaper]{article} 
\usepackage{aaai24}  
\usepackage{times}  
\usepackage{helvet}  
\usepackage{courier}  
\usepackage[hyphens]{url}  
\usepackage{graphicx} 
\usepackage{natbib}  
\usepackage{caption} 
\usepackage{algorithm}
\usepackage{algorithmic}
\usepackage{booktabs}
\usepackage{multirow}
\usepackage{colortbl}
\usepackage{xcolor}
\usepackage{threeparttable}
\usepackage{subfigure}
\usepackage{amssymb}
\usepackage{amsmath,amsfonts}
\usepackage{newfloat}
\usepackage{listings}
\DeclareCaptionStyle{ruled}{labelfont=normalfont,labelsep=colon,strut=off} 
\floatstyle{ruled}
\newfloat{listing}{tb}{lst}{}
\floatname{listing}{Listing}
\title{Learning Hierarchical Prompt with Structured Linguistic Knowledge for Vision-Language Models}
\author {
    Yubin Wang\textsuperscript{\rm 1},
    Xinyang Jiang\textsuperscript{\rm 2},
    De Cheng\textsuperscript{\rm 3},
    Dongsheng Li\textsuperscript{\rm 2},
    Cairong Zhao\protect\thanks{Corresponding Author (zhaocairong@tongji.edu.cn).}\textsuperscript{\rm 1}
}
\affiliations {
    \textsuperscript{\rm 1}Tongji University\\
    \textsuperscript{\rm 2}Microsoft Research Asia\\
    \textsuperscript{\rm 3}Xidian University\\
}

\usepackage{bibentry}

\begin{document}

\maketitle

\begin{abstract}

Prompt learning has become a prevalent strategy for adapting vision-language foundation models to downstream tasks. As large language models (LLMs) have emerged, recent studies have explored the use of category-related descriptions as input to enhance prompt effectiveness. Nevertheless, conventional descriptions fall short of structured information that effectively represents the interconnections among entities or attributes linked to a particular category. 
To address this limitation and prioritize harnessing structured knowledge, this paper advocates for leveraging LLMs to build a graph for each description to model the entities and attributes describing the category, as well as their correlations. Preexisting prompt tuning methods exhibit inadequacies in managing this structured knowledge. Consequently, we propose a novel approach called Hierarchical Prompt Tuning (HPT), which enables simultaneous modeling of both structured and conventional linguistic knowledge. 
Specifically, we introduce a relationship-guided attention module to capture pair-wise associations among entities and attributes for low-level prompt learning. 
In addition, by incorporating high-level and global-level prompts modeling overall semantics, the proposed hierarchical structure forges cross-level interlinks and empowers the model to handle more complex and long-term relationships.
Extensive experiments demonstrate that our HPT shows strong effectiveness and generalizes much better than existing SOTA methods. Our code is available at \url{https://github.com/Vill-Lab/2024-AAAI-HPT}.
\end{abstract}

\section{Introduction}
\begin{figure}[t]
    \centering
    \center{\includegraphics[width=7.9cm]{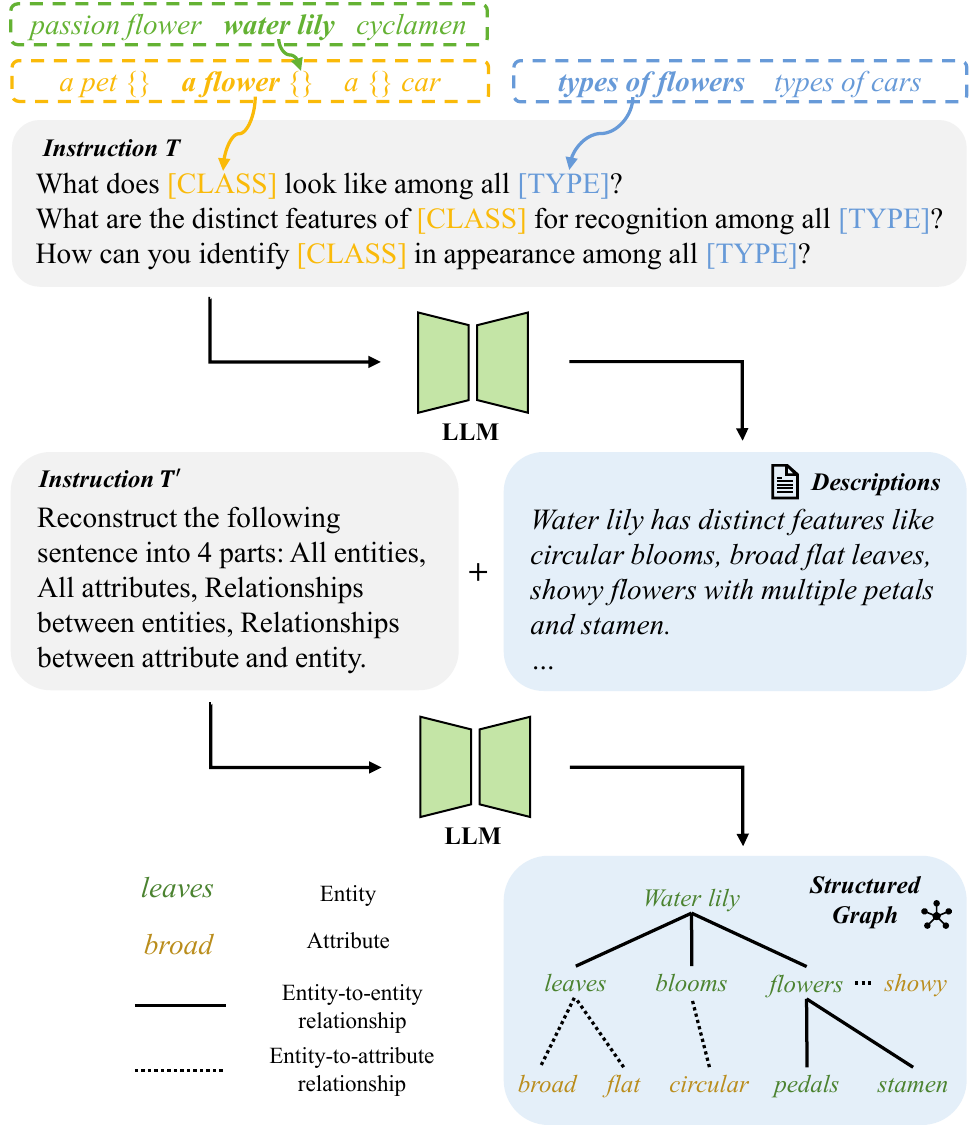}}
    \caption{We input a few hand-written instructions into LLM to generate human-like category-related descriptions along with structured graphs based on each description. \protect\label{fig:intro}}
\end{figure} 
Vision-Language foundation models (VLMs)~\cite{radford2021learning, jia2021scaling}, trained on large-scale datasets of image-text pairs, have made remarkable advancements in learning transferable representations. 
To effectively explore the potential of these powerful foundation models, prompt tuning methods~\cite{zhou2022learning, zhou2022conditional, khattak2023maple} aim to learn a set of continuous vectors known as prompt vectors and incorporate them in the input space, endowing the pre-trained network with a powerful representation capability. 
However, when confronted with ambiguous category names, models frequently struggle to make accurate judgments regarding the corresponding visual concepts, leading to underwhelming performance. 
Therefore, utilizing category names as text input without the assistance of linguistic knowledge seems to be a suboptimal choice. 
Recent methods~\cite{zhang2023prompt, pratt2022does, menon2022visual} have addressed this issue by using large language models (LLMs), such as GPT-3~\cite{brown2020language}. They take hand-written templates as input and generate human-like texts, containing rich linguistic knowledge that complements few-shot visual recognition. 

In this paper, we propose a novel approach to complement natural linguistic descriptions with a structured representation of knowledge.
We assert that this structured knowledge is essential for prompt tuning.
Specifically, the descriptions of a category with unstructured knowledge consist of key entities and attributes that define the category. For example, the category `water lily' is defined by entities like `leaves', `blooms', `flowers', each linked to category-specific attributes. 
Following related works on knowledge graphs~\cite{tay2017multi, zhang2021ma}, we represent these entities, attributes, and their correlations as a graph for semantic understanding.
This graph-based representation offers a more organized way to present information, leading to improved data comprehension. It facilitates the discovery of implicit connections that may not be evident in original descriptions.
In this work, we leverage existing large language models to obtain the structured information from vanilla descriptions, as shown in Figure \ref{fig:intro}.
Given a specific category, we feed hand-crafted instructions into LLMs, intending to generate human-like descriptions, as well as structured relationships within each description, including entities, attributes, and relationships among them. 

However, existing prompt tuning methods are inadequate to explicitly model the structured knowledge represented in a graph.
To this end, we propose \textbf{H}ierarchical \textbf{P}rompt \textbf{T}uning (HPT) to incorporate both structured and conventional linguistic knowledge from LLMs for enhancing prompt effectiveness in a hierarchical manner. 
To model the complex structured information, HPT learns hierarchical prompts with different semantic levels. 
Specifically, HPT contains low-level prompts representing the entities and attributes, high-level prompts with category-related information derived from descriptions, and global-level prompts with category-agnostic knowledge shared across categories.

To capture the LLM-generated pair-wise correspondences among entities and attributes, we introduce a relationship-guided attention module, where learnable attention-based matrices are integrated into the text encoder.
Furthermore, to handle more complex and long-term relationships not fully exploited by LLMs, cross-level self-attention is adopted to model relationships between prompts from different levels. It effectively overcomes the limitations caused by relying solely on the modeling of low-level tokens and allowing for a more comprehensive understanding of the category.
Our prompts are trained under a dual-path asymmetric framework~\cite{zhao2022learning}, where the prompted image encoder and text encoder are learned separately by aligning the output with the frozen encoder from the other modality respectively. 
By replacing the vanilla-prompted text encoder, which learns only category-agnostic prompts, with a novel hierarchical prompted text encoder, text representations can be better aligned with corresponding visual concepts, leading to excellent recognition performance.

The contributions of our work are summarized as follows. 
1) We raise the consideration that it is crucial to use structured knowledge from descriptions to assist learning prompts. 
Thus, we leverage large language models to generate category-related descriptions along with corresponding structured relationships; 
2) We propose Hierarchical Prompt Tuning (HPT) for simultaneously modeling both structured and conventional linguistic knowledge. 
By incorporating both forms of knowledge, we can enhance prompt effectiveness with more category-related information; 
3) Extensive experiments on three commonly used evaluation settings demonstrate remarkable improvements with our method.

\section{Related Work}

\subsection{Large Language Models}
Large Language Models (LLMs), such as GPT-3~\cite{brown2020language}, OPT~\cite{zhang2022opt}, and PaLM~\cite{chowdhery2022palm}, are trained on extensive web-scale datasets. 
Recently, ChatGPT~\cite{openai2023gpt4} has gained widespread popularity due to its strong ability to generate text resembling human-like writing and discern intricate patterns across diverse domains. Taking advantage of the vast potential of LLMs, recent studies have demonstrated their effectiveness in addressing various vision-language tasks~\cite{chen2022visualgpt, alayrac2022flamingo, yang2022zero}. 
Additionally, other studies investigate prompting vision-language models~\cite{zhang2023prompt, li2022bridge, wang2022learning} with LLMs for image classification, continuous learning, image caption generation, and action understanding. 
In this study, we aim to leverage the capabilities of LLMs in the field of the image classification task. 
When prompted with the target category, LLMs are able to generate related descriptions as well as corresponding structured relationships. 

\subsection{Visual-Language Models}
Large visual-language models (VLMs) have been instrumental in driving open vocabulary image classification, with CLIP~\cite{radford2021learning} being the pioneering work in this domain. 
Notable approaches include scaling up the models by using larger amounts of data, larger batch sizes, and bigger models, such as Align~\cite{jia2021scaling} and Basic~\cite{pham2021combined}, refining objective functions with models like SLIP~\cite{mu2022slip}, FILIP~\cite{yao2021filip}, and Lion~\cite{chen2023symbolic}, and incorporating supplementary information during training through models such as Florence~\cite{yuan2021florence}, UniCL~\cite{yang2022unified}, K-LITE~\cite{shen2022k}, and REACT~\cite{liu2023learning}. 
Our study is motivated by the desire to enhance the capabilities of CLIP with improved multi-modal prompts.

\subsection{Prompt Learning for V-L Models}
Prompt learning has its roots in natural language processing (NLP) and aims to enhance interaction with large language models~\cite{liu2023pre, brown2020language, wei2022chain}.  
Certain endeavors~\cite{menon2022visual, pratt2022does} propose leveraging pre-trained linguistic knowledge from LLMs to generate prompts, thereby enhancing V-L models without requiring additional training or labeling. 
To automate prompt engineering and explore optimal prompts, other studies~\cite{rao2022denseclip, zhou2022learning, zhou2022conditional, lu2022prompt} employ learnable text inputs and optimize them during training, known as prompt tuning. With the emergency of visual prompt tuning (VPT)~\cite{jia2022visual}, recent methods~\cite{khattak2023maple, zhao2022learning} take a multi-modal approach applying prompting on both modalities to improve alignment between vision and language representations. 
In contrast to prior studies, we generate diverse forms of linguistic knowledge and conduct hierarchical prompt tuning based on them to generate more robust representations.

\begin{figure*}[t]  
\begin{center}  
\subfigure[Overall pipeline for hierarchical prompt tuning]{  
\includegraphics[width=0.45\linewidth]{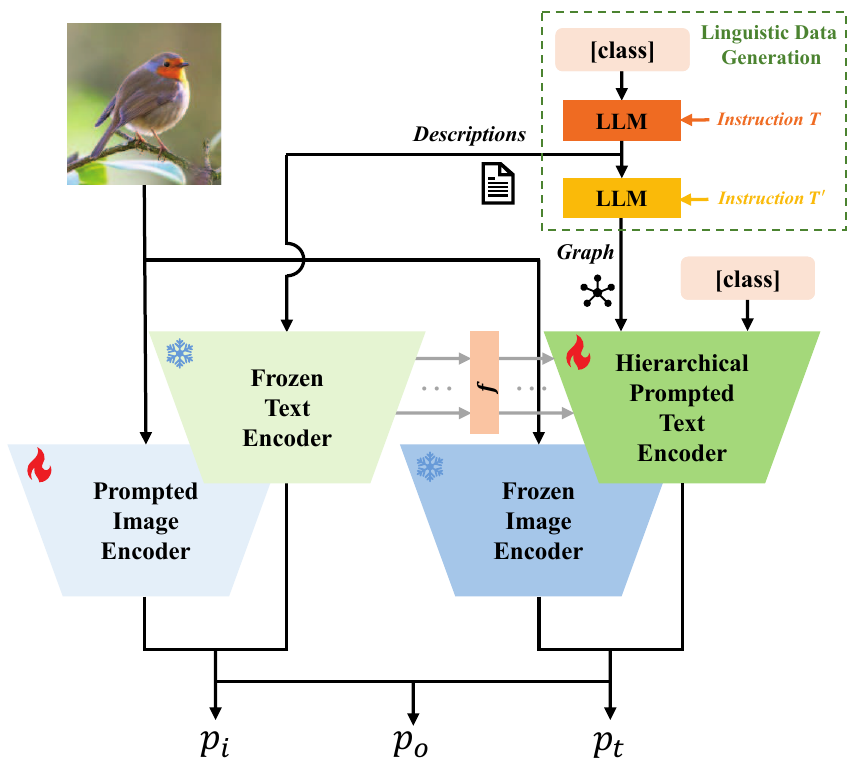}}\quad\quad
\subfigure[Structure of hierarchical prompted text encoder]{  
\includegraphics[width=0.46\linewidth]{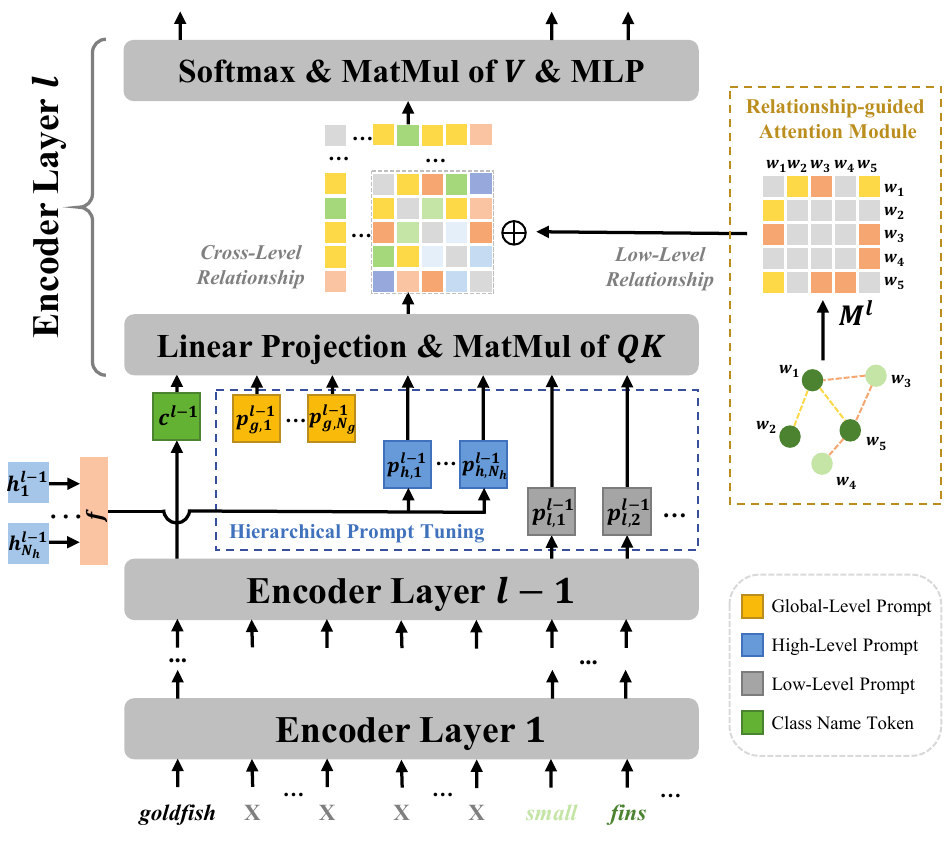}}
\caption{Our HPT applies a dual-path asymmetric network as the framework. Descriptions and relationship-guided graphs with class names are used as input for the frozen text encoder and the hierarchical prompted text encoder respectively. In the hierarchical prompted text encoder, we apply three types of prompts, low-level prompts, high-level prompts, and global-level prompts for hierarchical tuning, and design a relationship-guided attention module for better modeling structure knowledge. \protect\label{fig:framework}}  
\end{center}  
\end{figure*}

\section{Methodology}
\subsection{Overall Pipeline}
In this subsection, we will present the overall pipeline of our proposed method, as shown in Figure \ref{fig:framework}(a). In the context of a specific category, we initially input it with a set of hand-crafted templates as instruction into LLMs to generate human-like descriptions. 
Moreover, we further feed generated descriptions with another instruction into LLMs, aiming to capture the well-organized structure within each description, encompassing entities, attributes, and their relationships. We will provide a more detailed exposition in Section \textbf{Linguistic Data Generation}.

Given generated data, we apply a dual-path asymmetric network~\cite{zhao2022learning} for prompt tuning with visual-language models. This network experts in addressing overfitting issues associated with learned prompts, particularly in a few-shot learning scenario.
To conduct prompt tuning for transformer-like encoders, learnable vectors are introduced at each transformer layer’s input space as prompts.
The framework incorporates a novel asymmetric contrastive loss, which trains the prompted image encoder and text encoder separately with the frozen encoder from the opposite modality as guidance. 
Specifically, representations of prompted and frozen encoders from different modalities are aligned in an asymmetric way, leading to generating two probabilities $p_i$ and $p_t$ from the two frozen-prompted pairs. They are then averaged to derive an overall prediction $p_o$. 

Rather than making any modifications to visual prompts, we will mainly focus on prompt tuning for the text modality. In contrast to the prior dual-path asymmetric network, wherein two text encoders process identical text inputs, our approach adopts a distinct strategy that the frozen and prompted text encoders take entirely different inputs. In particular, unstructured descriptions are fed into the frozen encoder, while relationship-guided graphs along with the corresponding category name are fed into the novel hierarchical prompted encoder, which is specifically designed and finetuned for modeling structured information. 
In Section \textbf{Hierarchical Prompt Tuning}, we will dive into the core structure of this encoder for more details of tuning prompts from different semantic levels. 
To effectively capture the LLM-generated pair-wise correspondences among entities and attributes, the hierarchical prompted text encoder integrates a relationship-guided attention module, whose detailed implementation will be elaborated in Section \textbf{Relationship-guided Attention Module}.

\subsection{Linguistic Data Generation}
To acquire linguistic knowledge, we use one of the most powerful LLMs, ChatGPT~\cite{openai2023gpt4}, to generate descriptions with corresponding structured relationships. As shown in Figure \ref{fig:intro}, we adopt $N_h$ question templates as the language instruction $T$ for LLMs, e.g., ``What does a [CLASS] look like among all a [TYPE]?" or ``What are the distinct features of [CLASS] for recognition among all [TYPE]?", etc. [CLASS] denotes a specific category name with a modifier, like ``a pet Abyssinian". [TYPE] indicates the type of objects related to the dataset, like ``types of pets" for OxfordPets~\cite{parkhi2012cats}. We denote the generated descriptions from $T$ as $D=\{d_i\}_{i=1}^{N_h}$, formulated as

\begin{equation}
D = \operatorname{LLM}(T).
\end{equation}

For descriptions in $D$, we design an extra instruction $T^{'}$ to leverage LLMs for producing structured knowledge, including entities, attributes, and relationships among them. We denote the structured knowledge generated from $D$ as $R$, formulated as

\begin{equation}
R = \operatorname{LLM}([T^{'}, D]).
\end{equation}
Here $R=\{r_i\}_{i=1}^{N_h}$, $r_i=\{E_i, A_i, R_{e2e, i}, R_{e2a, i}\}$, where $E_i$, $A_i$, $R_{e2e, i}$, $R_{e2a, i}$ represent the entity set, the attribute set, the set of entity-entity relationships, and the set of entity-attribute relationships based on description $d_i$.

Our method utilizes both descriptions $D$ and structured knowledge $R$ as the source of category-related textual information, leading to effective prompt tuning.

\subsection{Hierarchical Prompt Tuning}
Given descriptions $D$ and structured knowledge $R$, we aspire to simultaneously model both structured and conventional linguistic knowledge. Therefore, we propose a novel approach called Hierarchical Prompt Tuning (HPT), which leverages both forms of knowledge for learning prompts in a hierarchical manner, as shown in Figure \ref{fig:framework}(b). HPT contains low-level prompts, high-level prompts, and global-level prompts, respectively denoted as $p_l$, $p_h$, $p_g$. 

\paragraph{Low-Level Prompt} To model pair-wise relationships within a description, we select essential words from this description as the input of the text encoder. Specifically, for entities in the entity set $E_i$ and attributes in the attribute set $A_i$, we simply concatenate them together as the low-level prompts $p_l^0$ for description $d_i$ and feed them into the first layer of the encoder. These prompts are seen as nodes in a relationship-guided graph, whose relationships are further processed by a novel relationship-guided attention module.

\paragraph{High-Level Prompt} In order to capture more intricate associations between individual tokens and the complete description, we derive high-level prompts $p_h$ that encapsulate the overall semantics of the category based on a series of descriptions. In detail, we feed descriptions $D$ into the frozen text encoder. Instead of simply utilizing representations from the last layer, we extract the last tokens from each layer containing rich semantics and feed them into a learnable prompt generator $f$, formulated as

\begin{equation}
p_{h,i}^l = f\left(h_i^l\right), 
\end{equation}
where $h_i^l$ represents the last token of description $d_i$ at the $l$-th layer. These tokens are then concatenated together as the high-level prompts $p_h^l=[p_{h,1}^l; ...; p_{h,{N_h}}^l ]$ of this category, which are further integrated into the corresponding layer of the hierarchical prompted encoder.

\paragraph{Global-Level Prompt} To represent category-shared knowledge pertinent to the task, we employ the standard approach for tuning the global-level prompts $p_g$. Instead of leveraging any form of knowledge, we automatically learn $N_g$ category-agnostic continuous vectors shared across categories as contexts and concatenate them with other prompts for each layer. 

\paragraph{Hierarchical Tuning} Based on the above prompts, we conduct the proposed hierarchical prompt tuning on the hierarchical prompted text encoder, formulated as
\begin{align} {\left[c^1, \_, \_, p_{l}^1\right] } & =L_1\left(\left[c, p_{g}^0, p_{h}^0, p_{l}^0\right]\right) \\ {\left[c^{i}, \_, \_, p_{l}^{i}\right]} & =L_i\left(\left[c^{i-1}, p_{g}^{i-1}, p_{h}^{i-1}, p_{l}^{i-1}\right]\right),\nonumber \\ i & = 2, 3, ..., N\end{align}
where $c$ represents the token of the class name. Via the projection head of the text encoder $\operatorname{TextProj}$, the final text representation $z$ is acquired by projecting the text embeddings $x^N$ corresponding to the last token of the last transformer block $L_N$ to a common V-L latent embedding
space,
\begin{equation}
z = \operatorname{TextProj} \left( x^N\right).
\end{equation}
\subsection{Relationship-guided Attention Module}
We introduce a relationship-guided attention module to model structured knowledge $R$ to capture pair-wise correspondences among entities and attributes in a layer-wise manner. 
For the $l$-th layer of a transformer-like encoder, an attention-based matrix $M^l$ is constructed based on generated relationships from each description. 
Two types of scalar values $\lambda_{e2e}^l$ and $\lambda_{e2a}^l$ are learned to indicate the strength of the relationship of entity-entity pairs and entity-attribute pairs separately. We assign the value to the respective element in the matrix, formulated as

\begin{equation}
M_{i,j}^l=\left\{
\begin{array}{ll}
\lambda_{e2e}^l & {(w_i, w_j) \in R_{e2e}}\\
\lambda_{e2a}^l & {(w_i, w_j) \in R_{e2a}}\\
0 & \mathrm{otherwise,}
\end{array}
\right.
\end{equation}
where $w_i$ indicates the entity or attribute associated with the $i$-th token in the sequence of low-level prompts.

Guided by structured knowledge, the learned attention-based matrices are integrated into layers of the text encoder. 
In practice, we compute the attention function on a set of queries simultaneously, packed together into a matrix $Q$. The keys and values are also packed together into matrices $K$ and $V$. For the $l$-th layer, with the attention-based matrix $M^l$, the output of self-attention is computed as
\begin{equation}
\operatorname{Attention}^l(Q, K, V)=\operatorname{softmax}\left(\frac{Q K^\top+M^l}{\sqrt{d_k}}\right) V.
\end{equation}
By explicitly adding $M^l$ into the calculation of self-attention, our model explicitly represents rich structured relationships within each description, thus enhancing crucial information associated with the category.

To deal with more intricate relationships, we include high-level and global-level prompts for the construction of long-term relationships. Unlike modeling correspondences with matrices, we automatically leverage the implicit associations through cross-level self-attention itself without any manual intervention. This design, as a hierarchical knowledge modeling approach, blends holistic semantics from multiple levels with structured relationships, thereby helping us discover complex associations that LLMs have failed to identify.

\section{Experimental Setup}
To evaluate our method, we follow the experiment setup established in previous works such as CoOp~\cite{zhou2022learning}, CoCoOp~\cite{zhou2022conditional}, and MaPLe~\cite{khattak2023maple}. We first describe evaluation protocols and datasets, followed by a discussion on implementation details.

\subsection{Evaluation Protocols}
\paragraph{Base-to-New Generalization} Aiming to evaluate the generalizability across various classes, this process involves dividing the dataset into base (seen) and new (unseen) classes and then training the model using a small number of samples from the base classes. Finally, we evaluate the model's performance on both base (few-shot performance) and new (zero-shot performance) classes. Additionally, we calculate the harmonic mean over the accuracy on both base and new classes to highlight the generalization trade-off.
\paragraph{Cross-Dataset Evaluation} This evaluation approach aims to assess the zero-shot ability of the model on a cross-dataset setup. To validate the potential of our approach in cross-dataset transfer, we train our model on all ImageNet classes in a few-shot manner and evaluate it directly on ten other unseen datasets with unknown categories in a zero-shot regime. 
\paragraph{Domain Generalization} To evaluate the robustness of our method on out-of-distribution datasets, we consider ImageNet as the source domain and its other variants as the target domain. We finetune our model on ImageNet in a few-shot setting and evaluate it on four variants of ImageNet with identical classes or subsets while manifesting diverse domain shifts.

\begin{table}[!b]
\centering
\resizebox{8.5cm}{!}{
\begin{threeparttable}
\begin{tabular}{lcccc|c|c}
\toprule 
Dataset & & CLIP & CoCoOp & MaPLe\tnote{*} & \textbf{HPT} & $\Delta$ \\
\midrule 
 & B & 69.34 & 80.47 & 82.28 & \textbf{84.32} & +2.04 \\
 \textit{Average} & N & 74.22 & 71.69 & 75.14 & \textbf{76.86} & +1.72\\
 & H & 71.70 & 75.83 & 78.55 & \textbf{80.23} & +1.68 \\
\midrule 
 & B & 72.43 & 75.98 & 76.66 & \textbf{77.95} & +1.29 \\
ImageNet & N & 68.14 & 70.43 & 70.54 & \textbf{70.74} & +0.20 \\
 & H & 70.22 & 73.10 & 73.47 & \textbf{74.17} & +0.70 \\
\midrule 
 & B & 96.84 & 97.96 & 97.74 & \textbf{98.37} & +0.41 \\
Caltech101 & N & 94.00 & 93.81 & 94.36 & \textbf{94.98} & +0.62 \\
 & H & 95.40 & 95.84 & 96.02 & \textbf{96.65} & +0.63 \\
\midrule 
 & B & 91.17 & 95.20 & 95.43 & \textbf{95.78} & +0.35 \\
OxfordPets & N & 97.26 & 97.69 & \textbf{97.76} & 97.65 & -0.11 \\
 & H & 94.12 & 96.43 & 96.58 & \textbf{96.71} & +0.13 \\
\midrule 
 & B & 63.37 & 70.49 & 72.94 & \textbf{76.95} & +4.01 \\
StanfordCars & N & 74.89 & 73.59 & 74.00 & \textbf{74.23} & +0.23 \\
 & H & 68.65 & 72.01 & 73.47 & \textbf{75.57} & +2.10 \\
\midrule 
 & B & 72.08 & 94.87 & 95.92 & \textbf{98.17} & +2.25 \\
Flowers102 & N & 77.80 & 71.75 & 72.46 & \textbf{78.37} & +0.57 \\
 & H & 74.83 & 81.71 & 82.56 & \textbf{87.16} & +4.60 \\
\midrule 
 & B & 90.10 & 90.70 & \textbf{90.71} & 90.46 & -0.25 \\
Food101 & N & 91.22 & 91.29 & \textbf{92.05} & 91.57 & -0.48 \\
 & H & 90.66 & 90.99 & \textbf{91.38} & 91.01 & -0.37 \\
\midrule 
 & B & 27.19 & 33.41 & 37.44 & \textbf{42.68} & +5.24 \\
FGVCAircraft & N & 36.29 & 23.71 & 35.61 & \textbf{38.13} & +1.84 \\
 & H & 31.09 & 27.74 & 36.50 & \textbf{40.28} & +3.78 \\
\midrule 
 & B & 69.36 & 79.74 & 80.82 & \textbf{82.57} & +1.75 \\
SUN397 & N & 75.35 & 76.86 & 78.70 & \textbf{79.26} & +0.56 \\
 & H & 72.23 & 78.27 & 79.75 & \textbf{80.88} & +1.13 \\
\midrule 
 & B & 53.24 & 77.01 & 80.36 & \textbf{83.84} & +3.48 \\
DTD & N & 59.90 & 56.00 & 59.18 & \textbf{63.33} & +3.43 \\
 & H & 56.37 & 64.85 & 68.16 & \textbf{72.16} & +4.00 \\
\midrule 
 & B & 56.48 & 87.49 & 94.07 & \textbf{94.24} & +0.17 \\
EuroSAT & N & 64.05 & 60.04 & 73.23 & \textbf{77.12} & +3.89 \\
 & H & 60.03 & 71.21 & 82.35 & \textbf{84.82} & +2.48 \\
\midrule 
 & B & 70.53 & 82.33 & 83.00 & \textbf{86.52} & +3.52 \\
UCF101 & N & 77.50 & 73.45 & 78.66 & \textbf{80.06} & +1.40 \\
 & H & 73.85 & 77.64 & 80.77 & \textbf{83.16} & +2.39 \\
\bottomrule
\end{tabular}
 \begin{tablenotes}
        \small
        \item[*]Previous SOTA method, the same for other generalization tasks.
      \end{tablenotes}
\end{threeparttable}
}
\caption{Comparison with existing methods on base-to-new generalization. B: Base Classes. N: New Classes. HM: Harmonic mean. $\Delta$: absolute improvement of HPT over the previous best result. HPT demonstrates strong generalization performance on 11 image recognition datasets. \protect\label{tab:b2n}}
\end{table}

\begin{table*}[!t]
\centering
\resizebox{17.7cm}{!}{
\begin{tabular}{lcccccccccccc}
\toprule
& \multicolumn{1}{c}{\textbf{Source}} & \multicolumn{11}{c}{\textbf{Target}} \\
\cmidrule(lr){2-2} \cmidrule(lr){3-13}
& ImNet & Caltech & Pets & Cars & Flowers & Food & Aircraft & SUN397 & DTD & EuroSAT & UCF & \textit{Average} \\
\midrule 
CoOp & 71.51 & 93.70 & 89.14 & 64.51 & 68.71 & 85.30 & 18.47 & 64.15 & 41.92 & 46.39 & 66.55 & 63.88 \\
CoCoOp & 71.02 & \textbf{94.43} & 90.14 & 65.32 & 71.88 & 86.06 & 22.94 & 67.36 & 45.73 & 45.37 & 68.21 & 65.74 \\
MaPLe & 70.72 & 93.53 & 90.49 & 65.57 & 72.23 & 86.20 & 24.74 & 67.01 & 46.49 & \textbf{48.06} & 68.69 & 66.30 \\
\midrule 
 \textbf{HPT} & \textbf{71.72} & 94.20 & \textbf{92.63} & \textbf{66.33} & \textbf{74.84} & \textbf{86.21} & \textbf{25.68} & \textbf{68.75} & \textbf{50.87} & 47.36 & \textbf{70.50} & \textbf{67.74} \\
\bottomrule
\end{tabular}}
\caption{Comparison with existing methods on cross-dataset evaluation. HPT achieves competitive performance providing the highest average accuracy, indicating superior generalization abilities on other datasets. \protect\label{tab:cde}}
\end{table*}

\subsection{Datasets}
For base-to-new generalization and cross-dataset evaluation, we follow the prior work~\cite{zhou2022learning} and evaluate the performance of our method on 11 image recognition datasets, which cover a wide range of recognition tasks. Specifically, the benchmark includes ImageNet~\cite{deng2009imagenet} and Caltech101~\cite{fei2004learning} for classification on generic objects; OxfordPets~\cite{parkhi2012cats}, StanfordCars~\cite{krause20133d}, Flowers102~\cite{nilsback2008automated}, Food101~\cite{bossard2014food} and FGVCAircraft~\cite{maji2013fine} for fine-grained classification; SUN397~\cite{xiao2010sun} for scene recognition; UCF101~\cite{soomro2012ucf101} for action recognition; DTD~\cite{cimpoi2014describing} for texture classification; and finally EuroSAT~\cite{helber2019eurosat} for satellite imagery recognition. For domain generalization, we utilize ImageNet as the source dataset and its four variants as target datasets including ImageNetV2~\cite{recht2019imagenet}, ImageNet-Sketch~\cite{wang2019learning}, ImageNet-A~\cite{hendrycks2021natural} and ImageNet-R~\cite{hendrycks2021many}.

\begin{table*}[!t]
\centering
\resizebox{13cm}{!}{
\begin{tabular}{lcccccc}
\toprule
& \multicolumn{1}{c}{\textbf{Source}} & \multicolumn{4}{c}{\textbf{Target}} \\
\cmidrule(lr){2-2} \cmidrule(lr){3-7}
& ImageNet & ImageNetV2 & ImageNet-S & ImageNet-A & ImageNet-R & \textit{Average} \\
\midrule 
CLIP & 66.73 & 60.83 & 46.15 & 47.77 & 73.96 & 57.17 \\
CoOp & 71.51 & 64.20 & 47.99 & 49.71 & 75.21 & 59.28 \\
CoCoOp & 71.02 & 64.07 & 48.75 & 50.63 & 76.18 & 59.90 \\
MaPLe & 70.72 & 64.07 & 49.15 & \textbf{50.90} & 76.98 & 60.26 \\
\midrule 
\textbf{HPT} & \textbf{71.72} & \textbf{65.25} & \textbf{49.36} & 50.85 & \textbf{77.38} & \textbf{60.71} \\
\bottomrule
\end{tabular}}
\caption{Comparison with existing methods on domain generalization. Overall, HPT shows consistent improvements on target variant datasets while achieving the highest accuracy on ImageNet. \protect\label{tab:dg}}
\end{table*}

\subsection{Implementation Details}
We apply prompt tuning to the pre-trained CLIP~\cite{radford2021learning} model, using ViT-B/16 as the visual backbone. We utilize SGD optimization with an initial learning rate of 0.0025 for base-to-new generalization and 0.001 for other tasks. Following the prior work~\cite{zhao2022learning}, the cross-entropy loss is adopted to equally minimize the discrepancy between the ground-truth label and the three aforementioned distributions $p_i$, $p_t$, $p_o$, while the overall distribution $p_o$ is used for inference. We randomly pick one description for each category to conduct relationship-guided attention learning during training for saving memory while leveraging all $N_h$ descriptions per category for inference. 

For base-to-new generalization, the maximum epoch is set to 10, with a batch size of 8. The length of global-level prompts $N_g$ is set to 2, and the number of descriptions for each category $N_h$, which is also the length of high-level prompts is set to 5. In accordance with the prior work~\cite{zhou2022learning}, we select 16 shots for training and the entire test set for evaluation. For domain generalization and cross-dataset evaluation, the maximum epoch is set to 3, with a batch size of 8, where we use the same hyperparameters for each dataset instead of a separate search.

\section{Experiments}
We evaluate our approach in three generalization settings, i.e. base-to-new generalization, cross-dataset evaluation, and domain generalization. We compare its performance with zero-shot CLIP~\cite{radford2021learning} and recent prompt learning works as strong baselines including CoOp~\cite{zhou2022learning} and CoCoOp~\cite{zhou2022conditional}, as well as the state-of-the-art method MaPLe~\cite{khattak2023maple}. In the case of CLIP, we use hand-crafted prompts specifically designed for each dataset. We further conduct several ablation experiments and sample analyses to better demonstrate the effectiveness of the proposed hierarchical prompt tuning.

\subsection{Base-to-New Generalization}
Table \ref{tab:b2n} presents the performance of HPT in base-to-new generalization setting on 11 recognition datasets.
Compared to the state-of-the-art prompt tuning method MaPLe, our approach achieves an enhancement of 1.72\% in terms of average accuracy for new classes, while simultaneously maintaining high accuracy on seen classes, even surpassing MaPLe by 2.04\%.
When considering both base and new classes, HPT shows an absolute average gain of 1.68\% on the harmonic mean over MaPLe, achieving a good trade-off between in-domain and out-of-domain data. 
The highest improvement of 4.64\% over the previous SOTA in the harmonic mean is observed for Flowers102.
With more available linguistic knowledge instead of only category names, our model trained by hierarchical prompt tuning shows a significant improvement.

\subsection{Cross-Dataset Evaluation}
Table \ref{tab:cde} shows the performance comparison between our HPT and existing methods on cross-dataset evaluation. On the ImageNet source dataset, HPT demonstrates comparable performance to competing approaches, yet it exhibits significantly superior generalization across 8 out of 10 datasets. Overall, HPT shows competitive performance leading to the highest average accuracy of 67.74\% with a gain of 1.44\% compared to the previous SOTA.  Unlike other methods that simply transfer the learned prompt vectors to new tasks, we provide a rich set of category-related knowledge as well as a novel hierarchical learning strategy for modeling the knowledge, leading to superior cross-domain performance.

\subsection{Domain Generalization}
We evaluate the direct transferability of our HPT trained on ImageNet to various out-of-domain datasets and observe that HPT consistently improves against all the existing approaches, as indicated in Table \ref{tab:dg}. Compared to MaPLe,
HPT performs slightly worse on ImageNet-A but better on the other three. As variant datasets share identical categories or subsets of categories with ImageNet, related linguistic knowledge from the source domain can be easily transferred, thereby assisting in recognizing out-of-domain data. 
 
\subsection{Ablation Experiments}
\paragraph{Influence of Different Prompts in HPT} We perform an ablation analysis on base-to-new generalization with various prompt combinations in HPT, as illustrated in Table \ref{tab:lvl}. The baseline method trains simply with global-level prompts. Experimental results show that both low-level and high-level prompts positively affect recognition performance. Among them, low-level prompts demonstrate a significant improvement in new classes, which shows the effectiveness of explicitly modeling structured relationships within descriptions thereby providing additional information linked to unfamiliar categories. High-level prompts also play an inseparable role in boosting performance by incorporating holistic semantics to handle more complex relationships. When all prompts are tuned with cross-level self-attention simultaneously, our model achieves optimal performance. 

\begin{table}[ht]
\centering
\begin{tabular}{ccc|cc|c}
    \toprule  
     \textbf{Global} & \textbf{High} & \textbf{Low} & \textbf{Base} & \textbf{New} & \textbf{HM}\\
    \midrule 
     $\checkmark$ &  &  & 84.02 & 75.20 & 78.99 \\
     $\checkmark$ & $\checkmark$ &  & 84.23 & 75.53 & 79.33 \\
     $\checkmark$ &  & $\checkmark$ & 84.05 & 76.11 & 79.59 \\
     $\checkmark$ & $\checkmark$ & $\checkmark$ & \textbf{84.32} & \textbf{76.86} & \textbf{80.23} \\
    \bottomrule 
\end{tabular} 
\caption{Ablation on different prompts in HPT. \protect\label{tab:lvl}}
\end{table}

\paragraph{Influence of Components in Relationship-guided Attention Module} As shown in Table \ref{tab:raa}, we perform an ablation study on combinations of components in the relationship-guided attention module, including entities and attributes, along with their relationships. Entities and attributes contribute essential insights extracted from descriptions indicating pertinent information. Consequently, they play an important role in aligning category-related text with corresponding visual concepts. Furthermore, by incorporating relationships that capture pair-wise correspondences among entities and attributes, we comprehensively model structured knowledge with vital information linked to the category, thereby leading to additional performance enhancements.

\begin{table}[ht]
\centering
\begin{tabular}{ccc|cc|c}
    \toprule  
    \textbf{Ent.} & \textbf{Attr.} & \textbf{Rel.} & \textbf{Base} & \textbf{New} & \textbf{HM}\\
    \midrule 
     &  &  & 84.23 & 75.53 & 79.33 \\
    $\checkmark$ &  &  & 84.21 & 75.76 & 79.49 \\
     & $\checkmark$ &  & 84.25 & 75.86 & 79.56 \\
    $\checkmark$ &  & $\checkmark$ & \textbf{84.34} & 76.00 & 79.71 \\
    $\checkmark$ & $\checkmark$ &  & 84.11 & 76.43 & 79.85 \\
    $\checkmark$ & $\checkmark$ & $\checkmark$ & 84.32 & \textbf{76.86} & \textbf{80.23} \\
    \bottomrule 
\end{tabular} 
\caption{Ablation on entities, attributes and their relationships in relationship-guided attention module. \protect\label{tab:raa}}
\end{table}

\begin{figure}[ht]
    \centering
    \center{\includegraphics[width=7.5cm]{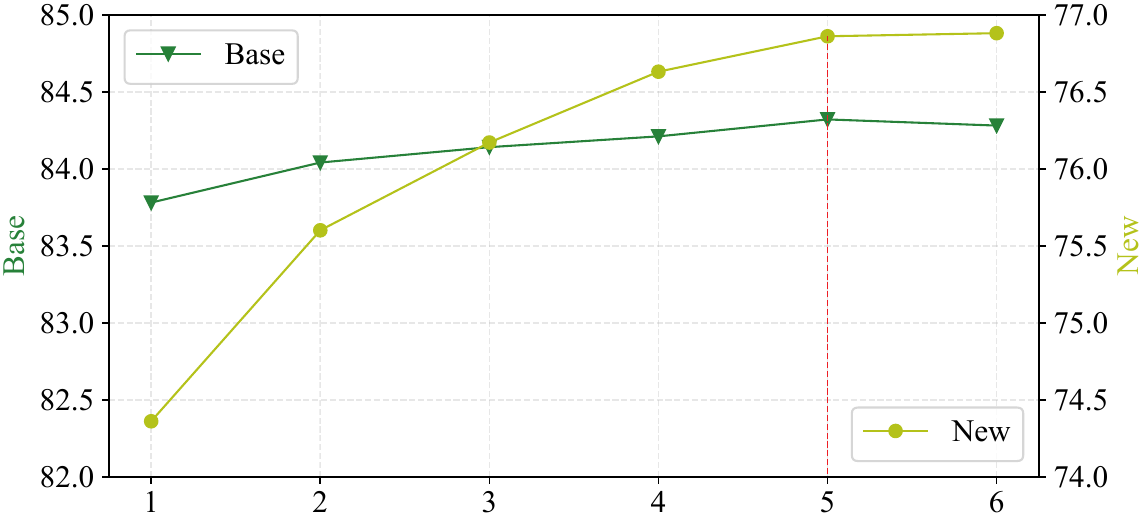}}
    \caption{Performance of HPT using different values of $N_h$. \protect\label{fig:nh}}
\end{figure}

\paragraph{Influence of the Number of Descriptions} We conduct experiments by varying the value $N_h$, the number of descriptions for each category. In Figure \ref{fig:nh}, as $N_h$ increases, the knowledge related to a category becomes richer, thus leading to consistent improvement in recognition accuracy. Notably, the impact on accuracy is considerably more pronounced for new classes compared to base classes. This is because, in the case of unseen classes where training images are unavailable, performance mainly relies on the diversity of linguistic knowledge. We set $N_h = 5$ for implementation as the accuracy barely changes when more information is provided. 

\subsection{Sample Analysis}
\begin{figure}[ht]
    \centering
    \center{\includegraphics[width=7.7cm]{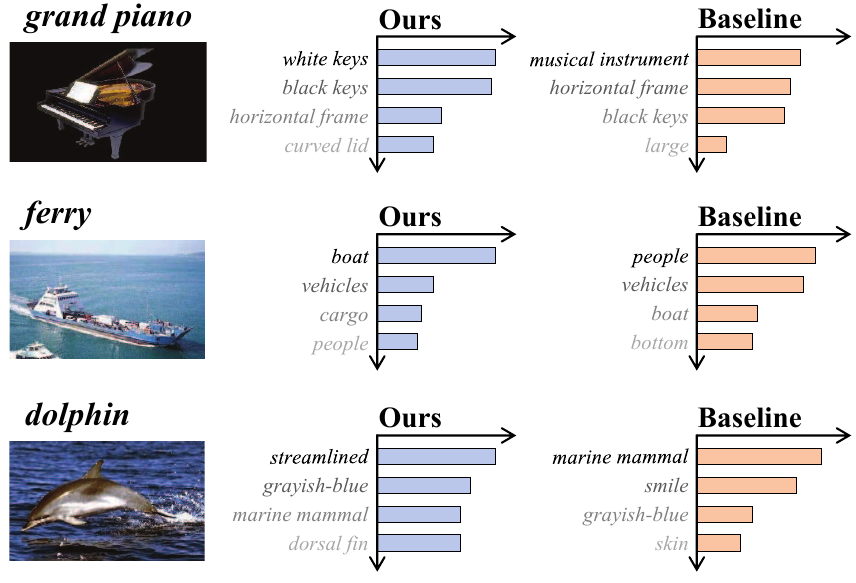}}
    \caption{Visualization of the top features with the highest attention scores according to the selected categories. \protect\label{fig:vis}}
\end{figure}

In order to demonstrate the capability of HPT to capture category-related semantics, we provide sample analysis on three randomly selected categories from Caltech101. Figure \ref{fig:vis} presents a comparison between our method and the baseline trained with the global-level prompts only. We observe the attention scores between tokens of entities and attributes from descriptions and the last token at the last layer of the prompted encoder. The top four features with the highest scores are displayed. It proves that HPT is capable of identifying discriminative visual concepts that significantly contribute to image recognition, leading to a substantial enhancement in the quality of text representations.

\section{Conclusion}
In this paper, we posit that utilizing structured relationships from descriptions to aid learning prompts is crucial. 
Consequently, we produce human-like descriptions accompanied by their corresponding structured relationships and present Hierarchical Prompt Tuning (HPT), a method that concurrently models both structured and conventional linguistic knowledge to strongly enhance prompt effectiveness. Our method demonstrates superior performance across three generalization tasks. We aspire that this work will garner increased attention toward the role of structured knowledge in natural language for prompt tuning, enabling its application to diverse tasks beyond classification.

\section{Ethical Statement}
The integration of ChatGPT in studies carries ethical implications with broad social ramifications. It enables inclusive communication but raises concerns about misinformation and biases. Ethical considerations demand transparency, bias mitigation, and ongoing evaluation to harness its benefits responsibly.

\section{Acknowledgments}
This work was supported by National Natural Science Fund of China (62076184, 61976158, 61976160, 62076182, 62276190), in part by   Fundamental Research Funds for the Central Universities and State Key Laboratory of Integrated Services Networks (Xidian University), in part by Shanghai Innovation Action Project of Science and Technology (20511100700) and Shanghai Natural Science Foundation (22ZR1466700).

\bibliography{aaai24}

\end{document}